\documentclass{article}

\usepackage{arxiv}

\usepackage[utf8]{inputenc} 
\usepackage[T1]{fontenc}    
\usepackage{hyperref}       
\usepackage{url}            
\usepackage{booktabs}       
\usepackage{amsfonts}       
\usepackage{nicefrac}       
\usepackage{microtype}      
\usepackage{lipsum}
\usepackage{graphicx}
\usepackage{amsmath}
\graphicspath{ {./images/} }

\title{Robustness and Resilience Evaluation of Eco-Driving Strategies at Signalized Intersections}

\author{
 Zhaohui Liang \\
  School of Engineering\\
  University of Wisconsin - Madison\\
  \texttt{zliang97@wisc.edu} \\
   \And
 Chengyuan Ma \\
  School of Engineering\\
  University of Wisconsin - Madison\\
  \texttt{cma97@wisc.edu} \\
  \And
 Keke Long \\
  School of Engineering\\
  University of Wisconsin - Madison\\
  \texttt{klong23@wisc.edu} \\
  \And
  Hang Zhou \\
  School of Engineering\\
  University of Wisconsin - Madison\\
  \texttt{hzhou364@wisc.edu} \\
 \And 
 Xiaopeng Li\\
 School of Engineering\\
 University of Wisconsin - Madison\\
 \texttt{xli2485@wisc.edu} \\
}

\begin{document}
\maketitle
\begin{abstract}
Eco-driving strategies have demonstrated substantial potential for improving energy efficiency and reducing emissions, especially at signalized intersections. However, evaluations of eco-driving methods typically rely on simplified simulation or experimental conditions, where certain assumptions are made to manage complexity and experimental control. This study introduces a unified framework to evaluate eco-driving strategies through the lens of two complementary criteria: control robustness and environmental resilience. We define formal indicators that quantify performance degradation caused by internal execution variability and external environmental disturbances, respectively. These indicators are then applied to assess multiple eco-driving controllers through real-world vehicle experiments. The results reveal key tradeoffs between tracking accuracy and adaptability, showing that optimization-based controllers offer more consistent performance across varying disturbance levels, while analytical controllers may perform comparably under nominal conditions but exhibit greater sensitivity to execution and timing variability.
\end{abstract}


\section{Introduction}
Eco-driving has long attracted attention as a promising strategy to reduce energy consumption and emissions \cite{jiang2025environmental}. This is particularly true in signalized intersections, where eco-driving can help avoid unnecessary stops and idling. Prior studies have shown that optimized intersection passing can significantly reduce fuel consumption and carbon emissions \cite{ma2021trajectory,reddy2024energy,mensing2014eco}. With the rapid advancement of connected and automated vehicle (CAV) technologies, vehicles now have access to precise traffic information and the ability to automatically execute planned actions, providing a practical foundation for large-scale deployment of eco-driving strategies.

CAV technologies have enabled the implementation of eco-driving strategies that combine trajectory planning with automated control. Most existing eco-driving strategies employ a hierarchical structure, in which an upper-level planning module generates the energy-efficient trajectory, and a lower-level controller ensures its execution. At the planning level, existing research has proposed various trajectory planning methods based on signal phase and timing (SPaT) information \cite{Jiang2017EcoAA,Sun2022AnEA}, queue information \cite{Li2022IndividualVS,yang2020eco}, and traffic flow predictions \cite{Long2024TrafficOM}. At the control level, advanced control algorithms, including model predictive control and physics-enhanced residual learning control \cite{dehkordi2019ecological}, have been used to enhance tracking performance. These efforts have demonstrated significant gains in fuel economy, stop reduction, and travel smoothness under ideal conditions.

 Existing eco-driving studies often evaluate performance through either microscopic traffic simulations or small-scale field experiments. While simulations allow for rapid iteration and control over input variables, they frequently abstract away key physical and implementation-level factors. In many simulation-based studies, the eco-driving strategy is validated solely at the planning level—where energy-optimal trajectories are generated and evaluated using pre-defined vehicle models—without incorporating a realistic control layer. As a result, execution-related issues such as actuator delays, tracking errors, and computational latency are ignored, making it impossible to assess the system's robustness in practice. On the other end, field tests can capture the full feedback loop between planning and control, but most existing implementations still operate under carefully designed, low-noise conditions. For instance, field validations are often limited to flat roads with fixed signal timings, no other traffic \cite{chen2016development,almannaa2019field,8095005} or low traffic density \cite{7447611}. Few studies explicitly quantify how system performance degrades under realistic disturbances, and even fewer report metrics that directly reflect execution fidelity or resilience to environmental changes.

In real-world deployment, however, eco-driving systems inevitably face both internal and external uncertainties. Internally, the nonlinear dynamics of vehicle actuators, time delays in command execution, and sensor inaccuracies can all lead to deviation from planned trajectories. These issues directly challenge the robustness of the control layer. For instance, \cite{zhao2024safe} reported degraded tracking performance due to actuator delays; \cite{mousa2019developing} observed unstable energy savings when signal plans changed unexpectedly; and \cite{wang2024sheeo} highlighted the limits of embedded systems in handling real-time optimization. Moreover, \cite{liang2024testing} demonstrated that Vehicle-to-Everything (V2X) communication delays and reduced packet reliability can significantly impair intersection control performance, aggravating throughput variance and safety-critical timing errors. On the other hand, the highly dynamic traffic environment, including signal changes \cite{8430781} and interactions with surrounding vehicles, requires the eco-driving planner to frequently re-adapt its trajectory—raising the need for resilience in the face of environmental disturbances. These \textbf{inherent uncertainties in the control system} and \textbf{variations in the external environment} impose unavoidable impacts on eco-driving performance, and should be explicitly considered in addition to the ideal-condition-based theoretical optimization.

To the best of our knowledge, existing eco-driving research has not yet provided a quantitative evaluation of such factors. In practice, the ability of an eco-driving system to withstand disturbances and instabilities largely determines the extent to which its advantages can be realized in real-world deployment. Therefore, in the study of eco-driving strategies for CAVs, it is essential to incorporate evaluation beyond efficiency-focused metrics such as fuel consumption, delay time, and number of stops, and to explicitly assess the system's robustness and resilience under uncertainty. 
To address these gaps, this study proposes robustness and resilience evaluation metrics for CAV eco-driving strategies at intersections. These metrics aim to extend current evaluation frameworks beyond pure efficiency, enabling a more grounded assessment of method applicability in uncertain conditions. We further conduct real-world vehicle experiments to benchmark several existing eco-driving strategies in terms of their control robustness and environmental resilience, thereby validating the practical utility of the proposed evaluation system.

The main contributions of this paper are as follows:
\begin{itemize}
    \item First to propose dedicated robustness and resilience metrics for CAV eco-driving strategies at intersections, enabling the evaluation of system performance under non-ideal real-world conditions and promoting practical deployment.
    \item The proposed metrics are validated via real-world testing of multiple eco-driving strategies, demonstrating its effectiveness in distinguishing method performance beyond theoretical efficiency.
\end{itemize}

The remainder of this paper is organized as follows. Section~\ref{sec:model} defines robustness and resilience metrics under two types of uncertainties. Section~\ref{sec:experiment} describes the experimental design and analyzes the evaluation results. Section~\ref{sec:conclusion} concludes the paper and outlines future work.

\section{Model} \label{sec:model}

\subsection{Problem Description}

We consider a general form of CAV eco-driving system at signalized intersections. For a CAV, let $t_0$ and $T$ denote the time at which the vehicle enters and completely exits the signalized intersection, respectively. The time horizon is discretized with a fixed interval $\Delta t$, thus the time set is defined as  $\mathcal{T} := \{t_0, t_0 + \Delta t, t_0 + 2\Delta t, \dots, T\}$.
Let $s_t \in \mathcal{S}$ denote the CAV’s state at time $t \in \mathcal{T}$, including variables such as position, velocity, and acceleration. $\mathcal{S}$ denotes the space of feasible states. At each time step, the vehicle can obtain external information $o_t \in \mathcal{O}$ via connectivity and self perception, including SPaT and the states of surrounding vehicles. 

Given the ego CAV state $s_t$ and external observations $o_t$, the eco-driving planner computes an energy-optimal trajectory over a future horizon $H$. This planning process is executed at a fixed interval $\delta$, where $\delta < H$, allowing continuous adaptation to updated ego and environmental states. Notably, the planning interval $\Delta t$ and the execution interval $\delta$ are intentionally modeled differently to accommodate systems with varying operating frequencies. The set of all planning time steps is defined as $\mathcal{T}_p := \{t_0, t_0 + \delta, t_0 + 2\delta, \dots, T - \delta\}$. The planning process at each time $t \in \mathcal{T}_p$ is denoted as
\begin{equation}
    \mathcal{J}^*_t = F(s_t, o_t) = \{ s^*_\tau \in \mathcal{S} \mid t \leq \tau \leq t + H \}
\end{equation}
where the output $\mathcal{J}^*_t$ represents the planned energy-optimal trajectory, and $F(\cdot)$ is the planning function, which varies depending on the specific eco-driving planning method employed.

Once the planned trajectory $\mathcal{J}^*_t$ is generated, a lower-level controller attempts to execute it in real-time. The actually executed trajectory is denoted as
\begin{equation}
    \mathcal{J} = \{ s_\tau \in \mathcal{S} \mid \forall \tau \in \mathcal{T}\}
\end{equation}
The formulation above represents a general eco-driving planning and control problem at signalized intersections, and it is sufficiently expressive to cover the existing methods in the literature. 

The modeling does not rely on an explicit mathematical expression of each disturbance factor. Instead, various sources of perturbations—including internal factors such as communication delays and computational latency, as well as external conditions such as lane changes, surrounding vehicle behaviors, signal timing variations, or potential cyber-attacks on information exchange—can be naturally integrated into the framework. These elements manifest as deviations between the planned trajectory and the actually executed trajectory under a given method, compared to its ideal behavior.

\subsection{Execution Deviation and Robustness Characterization}

In real-world implementations, the executed trajectory often deviates from the planned one. This discrepancy arises from multiple sources. On one hand, due to actuator dynamics and road geometry (e.g., slope, curvature), the lower-level controller may not accurately follow the planned states. On the other hand, many upper-level planning models intentionally simplify or neglect physical constraints to reduce computational complexity, resulting in trajectories that are not fully executable in practice.

We denote the actual executed state at time $t\in \mathcal{T}$ as
\begin{equation}
    s_t = s^*_t + z_t,
\end{equation}
where $z_t $ is the control error at time $t$, and shares the same dimensionality as the state space $\mathcal{S}$. The sequence of deviations $\mathbf{Z}= \{ z_t\mid t \in \mathcal{T} \}$ reflects the cumulative execution inaccuracy over the planning horizon.

This deviation characterizes the system’s robustness—its ability to maintain performance under bounded internal uncertainties. For a given eco-driving method $F(\cdot)$, a scenario-specific control task defined by the inputs $\textbf{X}= \{ (s_t, o_t) \mid \forall t \in \mathcal{T}_p \}$, and the actual control deviation \textbf{Z}, we define a robustness indicator $R(F(\cdot),\textbf{X},\textbf{Z})$, where a higher value indicates stronger robustness (i.e., smaller deviation from the planned trajectory or better tracking accuracy). Different planning strategies may exhibit varying robustness under the same scenario; for example, those that overly simplify control constraints or heavily rely on prediction accuracy tend to perform worse in this evaluation.

Under varying operating conditions and vehicle behaviors, the control deviation \textbf{Z} becomes a random variable with an associated distribution $\mathbb{Z}$. Given a specific set of scenario inputs \textbf{X}, the robustness indicator $R$ for method $F(\cdot)$ is also a random variable. Its statistical properties—such as the expectation $\mathbb{E}[H(F(\cdot),\textbf{X},\textbf{Z})]$—can be used to assess the average robustness of a given method across diverse conditions.

\subsection{Environmental Variation and Resilience Characterization}

In addition to internal execution errors, the eco-driving system must also contend with external environmental variations. Specifically, the external observations received at the next planning cycle, $o_{t} $, may differ from $o_{t- \delta}$ in the overlapping time horizon due to signal timing updates, surrounding vehicle behaviors, or lane changes. Let this discrepancy be denoted as $\Delta o_t $, which captures the difference between the external inputs relevant to the planned horizon at two adjacent planning cycles.

Such variations may invalidate portions of the previously planned trajectory $\mathcal{J}^*_t$, necessitating online re-planning to maintain safety and energy efficiency. The collection of these deviations is represented by $\Delta\textbf{O} := \{ \Delta o_t \mid \forall t \in \mathcal{T}_p \}$. The degree to which the planned trajectory $\mathcal{J}^*_t$ is sensitive to $\Delta\textbf{O}$ reflects the system’s \textit{resilience}—its ability to adapt to environmental disturbances while preserving goal satisfaction and operational stability.

We define a resilience indicator $G(F(\cdot), \Delta\textbf{O})$, which evaluates how effectively the eco-driving strategy $F(\cdot)$ adapts its trajectory in response to a given disturbance $\Delta\textbf{O}$. A higher value of $G$ implies stronger resilience to the corresponding variation.

Across diverse driving conditions and environmental dynamics, $\Delta\textbf{O}$ follows a distribution $\mathbb{D}$. Consequently, $G(F(\cdot), \Delta\textbf{O})$ becomes a random variable with an associated distribution $\mathbb{G}$, and its expected value $\mathbb{E}_{\Delta \textbf{o} \sim \mathbb{D}} [G(F(\cdot), \Delta\textbf{O})]$ quantifies the average resilience of the method under realistic disturbances.

Therefore, the complete eco-driving process involves repeated computation of $F(s_t, o_t)$ to generate $\mathcal{J}_t$, which is subsequently executed as $\mathcal{J}$, subject to both internal and external uncertainties. These uncertainties jointly contribute to the gap between ideal energy-efficient behavior and its practical realization, highlighting the necessity of systematically evaluating both robustness and resilience.

\subsection{Model of the Robustness Indicator and the Resilience Indicator}

To quantify the performance impact of control deviations, we define a general utility function \( U(\mathcal{J}) \in \mathbb{R}^{+} \) that maps a given trajectory \( \mathcal{J} \subseteq \mathcal{S} \) to a negative scalar representing the overall control quality. This utility function can be flexibly specified to reflect key performance objectives, such as fuel consumption, velocity smoothness, or adherence to a reference trajectory. For optimization-based planning methods, the utility function \( U(\cdot) \) can directly correspond to the optimization objective. Without loss of generality, we assume that the utility function \( U(\cdot) \) satisfies the following properties:
\begin{itemize}
    \item[(1)] \textbf{Trajectory dependency:} \( U \) depends solely on the trajectory \( \mathcal{J} \), i.e., it is a functional of the state sequence and does not rely on other auxiliary variables.
    \item[(2)] \textbf{Temporal additivity:} \( U \) is temporally additive (i.e., interval-decomposable), meaning that for any partition \( t_1 < t_2 < t_3 \), the utility over the combined interval satisfies
    \[
        U\left( \mathcal{J}(t_1, t_3] \right) = U\left( \mathcal{J}(t_1, t_2] \right) + U\left( \mathcal{J}(t_2, t_3] \right).
    \]
    \item[(3)] \textbf{Monotonicity:} Larger values of \( U(\cdot) \) indicate better performance.
\end{itemize}

Under the assumption of no external disturbances (i.e., neglecting the effect of $\Delta\textbf{O}$), internal control errors $\textbf{Z}$ result in an executed trajectory denoted as \( \mathcal{J}^\mathcal{Z} \).

Since eco-driving strategies are typically optimized over conventional human driving behavior, there exists a performance baseline represented by a benchmark trajectory \( \mathcal{J}^B \) (e.g., from a rule-based or non-eco driving policy) over the same interval.

For a certain test, we define the robustness indicator as the relative performance retention ratio:

\begin{equation}
    R(F(\cdot)) = \frac{U(\mathcal{J^Z}) - U(\mathcal{J}^B)}{\sum_{t \in \mathcal{T}_p} \left[ U(\mathcal{J}^*_t{(t, t+\delta]}) - U(\mathcal{J}^B) \right]}
\end{equation}

In principle, \( R(F(\cdot)) \in [0,1] \) quantifies the proportion of the theoretically achievable performance gain (relative to the benchmark) that is retained under control execution errors. A value of \( R = 1 \) indicates that the executed trajectory fully preserves the expected benefit of the planned trajectory, whereas lower values signify greater degradation due to lack of robustness. However, in practice, it is also possible for \( R > 1 \) to occur, if certain disturbances accidentally lead to better-than-planned outcomes.

We now extend the formulation to account for the effect of external disturbances \( \Delta\textbf{O} \), which, together with internal control errors \( \textbf{Z} \), jointly determine the actual executed trajectory \( \mathcal{J} \). 

We define the resilience indicator as:
\begin{equation}
   G(F(\cdot)) = \frac{U(\mathcal{J}) - U(\mathcal{J}^B)}{\sum_{t \in \mathcal{T}_p} \left[ U\left( \mathcal{J}^*_t(t, t+\delta] \right) - U(\mathcal{J}^B) \right]}
\end{equation}
where \( \mathcal{J} \) reflects the final trajectory under the influence of both internal and external disturbances. A value of \( G = 1 \) indicates perfect adaptation to disturbances with no loss in performance, while lower values reflect increasing degradation caused by external variations. 

Since \( \Delta o_t \) follows a distribution \( \mathbb{D} \), the resilience indicator \( G(F(\cdot), \Delta\textbf{O}) \) becomes a random variable with an associated distribution \( \mathbb{G} \). Its expected value \( \mathbb{E}[G(F(\cdot), \Delta\textbf{O})] \) serves as a measure of the average resilience of the eco-driving method under real-world uncertainties.

This formulation enables a unified evaluation framework, where both robustness and resilience are expressed through normalized performance retention ratios under respective types of disturbances.

\section{Experiment}\label{sec:experiment}
\subsection{Experimental Setup and Test Scenario}

\begin{figure}[h]
    \centering
    \includegraphics[width=0.95\linewidth]{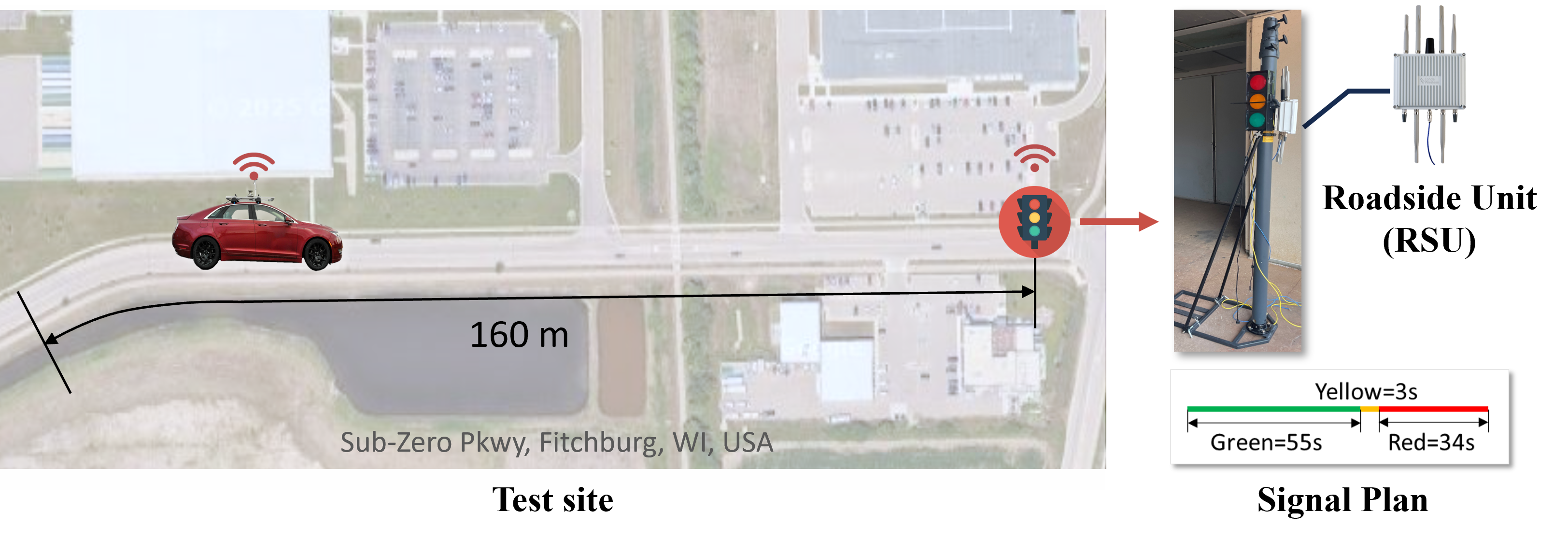}
    \caption{Real-world experimental setup.}
    \label{fig:test_scenario}
\end{figure}

To evaluate the proposed robustness and resilience metrics, we conduct real-world vehicle experiments at a signalized intersection. Figure~\ref{fig:test_scenario} illustrates the experimental setup, including the test site layout, communication infrastructure, and signal timing configuration.

The test site geometry and signal timing parameters are derived from the NGSIM dataset collected at Peachtree Street in Atlanta, GA \cite{NGSIM_Peachtree_2007}. The vehicle approaches the intersection from a minor road with an approach distance of 160 meters. To ensure consistent initial conditions across all tests, the CAV enters the test zone at a constant speed of 5 m/s and maintains this speed until the eco-driving controller begins trajectory optimization.

Real-world disturbances affecting CAVs can vary widely in both form and probability, including environmental variability, sensor noise, and unexpected signal changes. In a related study, we investigated adversarial scenario generation under diverse conditions to explore the boundaries of controller robustness. In this work, however, we focus on a controlled and repeatable disturbance scenario designed to evaluate the resilience of the controller under signal-related disruptions.

Specifically, we simulate an actuated signalized intersection where the CAV approaches from a minor road. Upon its arrival, the red phase of the traffic light is intentionally extended, creating a deterministic but adverse signal disturbance. This setup serves as a numerical case study to assess the controller's ability to maintain safety and performance under delayed signal clearance.

The intersection geometry and signal timing configuration are derived from...

Real-world disturbances affecting CAVs can vary widely in both form and probability, including environmental variability, sensor noise, and unexpected signal changes. In a related study, we investigated adversarial scenario generation under diverse conditions to explore the boundaries of controller robustness. In this work, however, we focus on a controlled and repeatable disturbance scenario designed to evaluate the resilience of the controller under signal-related disruptions.

Specifically, we simulate an actuated signalized intersection where the CAV approaches from a minor road. Upon its arrival, the red phase of the traffic light is intentionally extended, creating a deterministic but adverse signal disturbance. This setup serves as a numerical case study to assess the controller's ability to maintain safety and performance under delayed signal clearance.

The intersection geometry and signal timing configuration are derived from the NGSIM dataset collected at Peachtree Street in Atlanta, GA \cite{NGSIM_Peachtree_2007}. In this scenario, the vehicle starts from position 0, with the traffic light located at the 160-meter mark. The baseline signal timing is set to 34.8 seconds of red, 3.2 seconds of yellow, and 55.1 seconds of green. To ensure a well-posed arrival time calculation $t_p=\frac{Distance}{v_{t0}}$ , the CAV enters the intersection at a constant speed of 5 m/s.

According to the Manual on Uniform Traffic Control Devices (MUTCD) \cite{FHWA_MUTCD_2023}, actuated signal systems typically apply green time extensions ranging from 2 to 6 seconds, depending on prevailing traffic volume and approach speed. Guided by this standard, we incorporate green light extensions of 0, 2, 4, and 6 seconds into our simulations as structured disturbance inputs to evaluate the controller's resilience across varying levels of temporal disruption.

\subsection{Test platform}
Variations in computational capacity and vehicle actuator characteristics can lead to discrepancies in control performance, even when the same control algorithm is applied. To ensure fair and consistent evaluation in field experiments, it is crucial to minimize these sources of variability by standardizing the testing conditions and platform.

\begin{figure}
    \centering
    \includegraphics[width=0.75\linewidth]{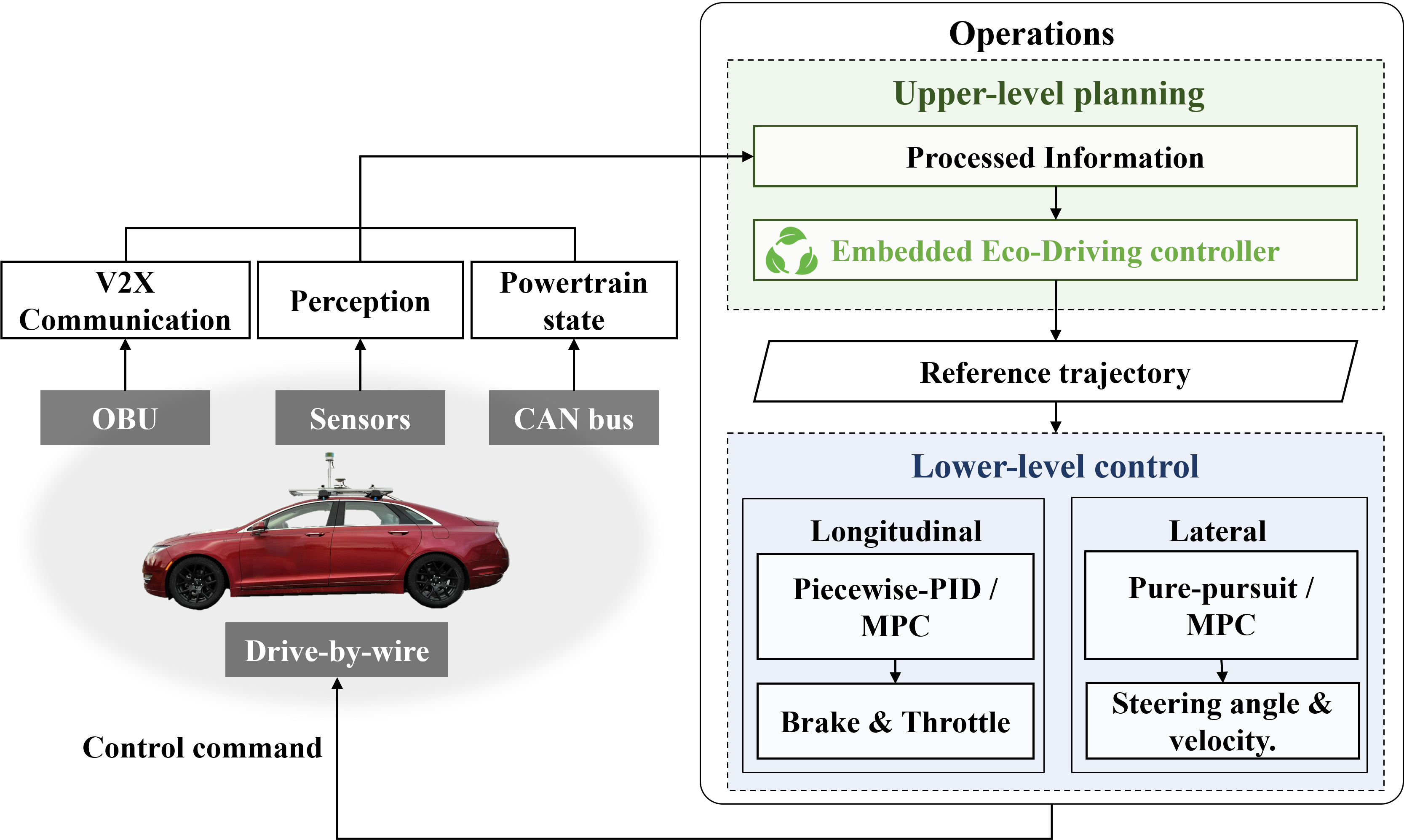}
    \caption{Test platform}
    \label{fig:platform}
\end{figure}

Figure \ref{fig:platform} illustrates the system architecture of the Level 3 autonomous vehicle (AV) and the flow of information across its modules. The perception module processes raw data from the onboard sensor suite and performs sensor fusion to produce actionable environmental information. Simultaneously, the onboard unit (OBU) receives communication packets from surrounding connected infrastructure and vehicles. Both modules feed structured data into the operation module, which is responsible for motion planning and control.

The operation module follows a hierarchical design. At the high level, the eco-driving controller is implemented as a modular plugin, enabling seamless switching between different control algorithms without affecting the rest of the system. This controller generates reference trajectories over a predictive horizon. At the low level, control commands—such as those for throttle, brake, and steering—are computed based on the reference trajectory and current vehicle states (e.g., drivetrain signals). These commands are then converted into CAN bus messages and transmitted to the drive-by-wire (DBW) system for execution.

\subsection{Embedded Eco-Driving Controllers }

\paragraph{Stop-and-Go Controller - Benchmark}
The \textit{Stop-and-Go} controller emulates human-like driving behavior. The CAV cruises at a desired constant speed $v_{\text{exp}}$ until it reaches the red light position $x_{\text{light}}$, where it comes to a complete stop and waits for the green signal. Let $[t_r^{\text{start}}, t_r^{\text{end}}]$ denote the red light time interval. The controller's behavior can be divided into three distinct phases:
\begin{itemize}
    \item \textbf{Cruising Phase:}
    \begin{align}
        \text{If } x_t < x_{\text{light}} \text{ and } t \in [t_r^{\text{start}}, t_r^{\text{end}}], \text{ then } v_t = v_{\text{exp}}
    \end{align}
    
    \item \textbf{Stopping Phase:}
    \begin{align}
        \text{If } x_t \geq x_{\text{light}} \text{ and } t \in [t_r^{\text{start}}, t_r^{\text{end}}], \text{ then } v_t = 0
    \end{align}
    
    \item \textbf{Go Phase:}
    \begin{align}
        \text{If } t > t_r^{\text{end}}, \text{ then } v_t = v_{\text{exp}}
    \end{align}
\end{itemize}

\paragraph{Optimization-Based Controller}
The \textit{optimization-based controller} aims to minimize fuel consumption by solving a trajectory optimization problem over a finite time horizon. The cost function is constructed using a regression-based fuel model, where $\alpha, \beta, \gamma, \delta, \epsilon$ are coefficients derived from empirical data \cite{shi2022empirical}, and $j_t$ denotes the jerk:
\begin{align}
   \min\quad&  \sum_{t=0}^{T} \left( \alpha + \beta v_t + \gamma v_t^2 + \delta v_t\cdot a_t + \epsilon a_t^2 \right) \\
   \text{s.t.} \quad & x_{t+1} = x_t + v_t \cdot \delta  + \frac{1}{2} a_t \cdot \delta ^2, & \forall t \in \mathcal{T} \setminus \{T\}, \\
& v_{t+1} = v_t + a_t \cdot \delta , & \forall t \in \mathcal{T} \setminus \{T\}, \\
& a_{t+1} = a_t + j_t \cdot \delta , & \forall t \in \mathcal{T} \setminus \{T\}, \\
& x_1 = x_0, \\ 
& v_1 = v_0, \\
& x_T = 0, \\
& v_T = v_p, \\
& v_{\min} \leq v_t \leq v_{\max}, & \forall t \in \mathcal{T}, \\
& a_{\min} \leq a_t \leq a_{\max}, & \forall t \in \mathcal{T}, \\
& j_{\min} \leq j_t \leq j_{\max}, & \forall t \in \mathcal{T}.
\end{align}

\paragraph{Analytical Trajectory Controller}

The \textit{analytical trajectory controller} generates a smooth speed profile using cubic polynomials, based on boundary conditions, without requiring iterative optimization. It handles three distinct cases depending on the timing relative to the signal phase. Let $t_e$, $t_p$, and $t_c$ denote the earliest possible arrival time, predicted arrival time, and a critical threshold time, respectively. Detailed derivation of these timings can be found in \cite{liang2025analytical}.
\begin{itemize}
    \item \textbf{Case 1: $t_e = t_p$} \\
    The CAV cruises at the expected speed $v_p$ to pass through the intersection.

    \item \textbf{Case 2: $t_e < t_p < t_c$} \\
    The vehicle follows a cubic trajectory defined by:
    \begin{align}
        x(t) = a t^3 + b t^2 + c t + d
    \end{align}
    The coefficients $a$, $b$, $c$, and $d$ are solved analytically using the initial and final boundary conditions.
    \item \textbf{Case 3: $t_p > t_c$} \\
    The vehicle decelerates along a cubic trajectory to a full stop at position $x_s$, waits for a duration $t_w$, and then resumes motion using another cubic polynomial:
    \begin{align}
        x(t) =
        \begin{cases}
        a t^3 + b t^2 + c t + d, & 0 \leq t < t_s \\
        x_s, & t_s \leq t < t_s + t_w \\
        a (t - t_w)^3 + b (t - t_w)^2 + c (t - t_w) + d, & t_s + t_w \leq t \leq t_c
        \end{cases}
    \end{align}
\end{itemize}

\paragraph{Implementation Strategy}
Both the optimization-based and analytical controllers are implemented using a \textit{rolling horizon approach}. At each time step, the controller replans the trajectory based on the current state and updated traffic signal information. This allows real-time adaptation to environmental changes and uncertainties.

\subsection{Test result}
To evaluate the deviation between the executed and planned trajectories, we adopt the classical root mean square error (RMSE) metric, which computes the average deviation between the actual state \( s_t \) and the planned state \( s^*_t \) over the entire execution horizon:
\begin{equation}
    \text{RMSE} = \sqrt{ \frac{1}{T} \sum_{t=0}^{T} \| s_t - s^*_t \|^2 }.
\end{equation}
This metric quantifies the average tracking error and serves as a baseline indicator for control accuracy.

In addition to RMSE, we evaluate the control robustness and environmental resilience of each method using the proposed indicators \( R \) and \( G \), respectively. The stop-and-go trajectory, which represents conventional conservative driving behavior, is used as the benchmark trajectory \( \mathcal{J}^B \) for normalization in both metrics.

The overall utility function \( U(\mathcal{J}) \) used in the robustness and resilience evaluation combines two objectives: (1) travel time efficiency, and (2) energy consumption estimated from the trajectory profile. A weighted sum formulation is adopted:
\begin{equation}
    U(\mathcal{J}) = w_1 \cdot \left(-T_{\text{pass}} \right) + w_2 \cdot \left( -E(\mathcal{J}) \right),
\end{equation}
where \( T_{\text{pass}} \) is the total time to pass through the intersection, \( E(\mathcal{J}) \) is the estimated energy consumption over the trajectory, and \( w_1, w_2 \) are weighting coefficients. Higher values of \( U(\cdot) \) represent better performance in terms of faster and more energy-efficient passing. 

Following \cite{shi2022empirical}, we model instantaneous fuel consumption as a quadratic polynomial in speed \(v\) and acceleration \(a\):
\begin{align}
    E(\mathcal{s_t})= \alpha + \beta v + \gamma v^2 + \theta v a + \eta a^2,
\end{align}
whose coefficients were $\alpha = 0.15,\beta = 0.0025,\gamma = 0.00006,\theta = 0.00035,\eta = 0.0004 $.

This evaluation framework allows for a comprehensive comparison of different eco-driving methods, capturing both traditional tracking accuracy and higher-level behavioral stability under disturbances.

\begin{table}[ht]
    \centering
    \caption{Resilience Evaluation of Different Eco-Driving Methods under Real-World Tests}
    \begin{tabular}{llccccc}
        \toprule
          Scenarios&Method & \( U(\mathcal{J}) \)& \( U(\mathcal{J}^B) \) & \(U(\mathcal{J}^*) \)& RMSE(m)& \( G \)  \\
          \midrule
         1&Method A (Optimization) &  -5.6440& -8.9725&  -5.7147&  1.0254& 1.0825 \\
         &Method B (Analytical)   &  -5.5764& -8.9725&  -5.6620&  1.3716& 1.0788 \\
         2& Method A (Optimization) 
        & -5.4588& -8.4446& -5.4710& 0.6622&1.0560 \\
         & Method B (Analytical)   & -5.7243& -8.4446& -5.7009& 1.1608&1.0320 \\
         3& Method A (Optimization) 
        & -5.7283& -8.6956& -5.7682& 0.7092&1.0699 \\
         & Method B (Analytical)   & -5.6609& -8.6956& -5.6569& 0.9362&1.0319 \\
         4& Method A (Optimization) 
        & -5.4048& -8.5720& -5.4291& 0.9306&1.0689 \\
         & Method B (Analytical)   & -5.5992& -8.5720& -5.6229& 1.2626&1.0689 \\
         \bottomrule
    \end{tabular}
    \label{tab:evaluation}
\end{table}

\begin{table}[ht]
    \centering
    \caption{Robustness Evaluation of Different Eco-Driving Methods under Real-World Tests}
    \begin{tabular}{ccccccc}
        \toprule
        Test& Method& $U(\mathcal{J}) $& $U(\mathcal{J}^B) $& $U(\mathcal{J}^*) $& RMSE(m)& $R$\\
        \midrule
 1& Method A (Optimization) 
& -5.3413& -8.8698& -5.4617& 1.2509&1.0354\\
        & Method B (Analytical)   & -5.6034& -8.8698& -5.6655& 0.7839& 1.0194\\
        2& Method A (Optimization) 
& -5.3732& -9.2524& -5.4439& 1.1373& 1.0182\\
 & Method B (Analytical)   & -5.4023& -9.2524& -5.4300& 2.4526&1.0072\\
        3& Method A (Optimization) 
& -6.2175& -8.7952& -6.2383& 0.6882& 1.0082\\
 & Method B (Analytical)   & -5.7234& -8.7952& -5.7707& 0.8783&1.0156\\
        \bottomrule
    \end{tabular}
    \label{tab:robust}
\end{table}

Table~\ref{tab:evaluation} summarizes the performance of two embedded eco-driving controllers—\textbf{Method A (Optimization-based)} and \textbf{Method B (Analytical)}—across four real-world scenarios with increasing disturbance severity. The evaluation includes five metrics: the utility of the executed trajectory $U(\mathcal{J})$, the utility of a fixed baseline benchmark $U(\mathcal{J}^B)$, the utility of the planned trajectory $U(\mathcal{J}^*)$, the root-mean-square tracking error (RMSE), and the resilience indicator $G$ as defined in Equation~\ref{equ:resilience indicator}. These metrics collectively quantify tracking performance, execution quality, and disturbance sensitivity.

The resilience indicator $G$ captures how much utility is lost due to execution errors and external disturbances, normalized by the ideal-to-baseline gap. Importantly, $G$ represents relative degradation—higher values imply greater divergence from planned utility, even if absolute performance remains strong. Across scenarios 1 to 4, both controllers experience increasing RMSE and utility degradation, consistent with more severe signal phase disturbances (e.g., longer red phase extensions). This validates the design of $G$ as a disturbance-sensitive performance indicator.

\begin{figure}
    \centering
    \includegraphics[width=1\linewidth]{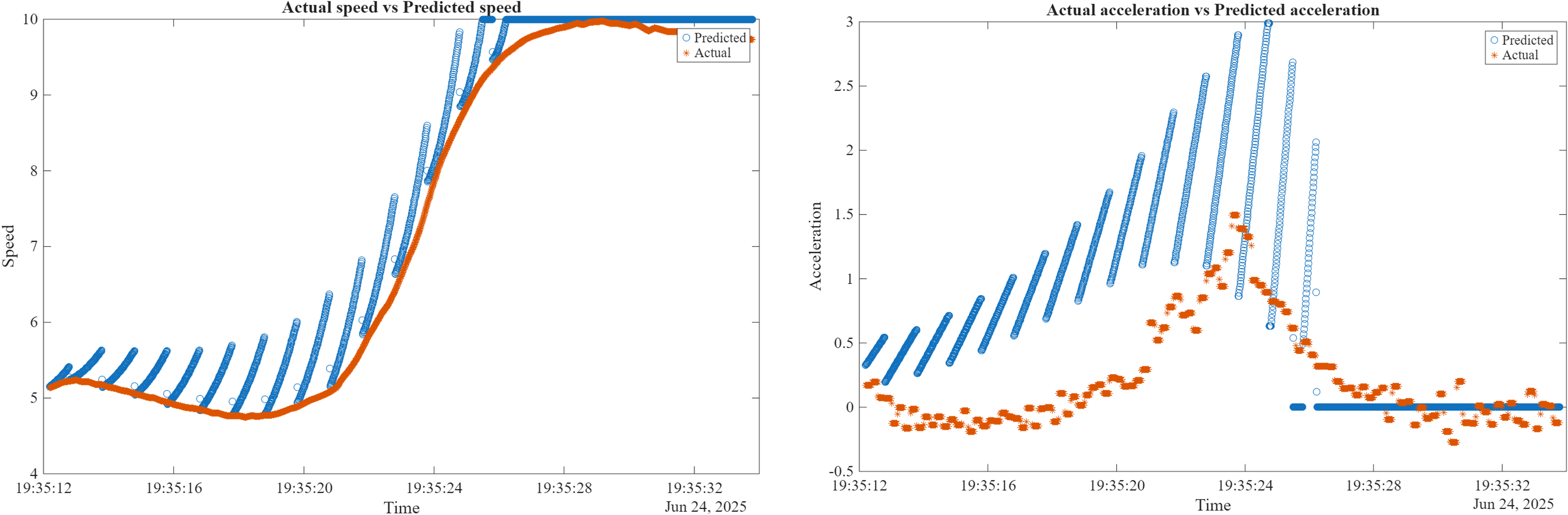}
    \caption{Actual vs Prediction}
    \label{fig:actvspre}
\end{figure}

An interesting pattern is that the executed trajectories often show higher utility than the planned ones. This phenomenon can be attributed to the lagging execution caused by internal delays and system inertia. As seen in Figure~\ref{fig:actvspre}, the controller fails to precisely follow the high-acceleration maneuvers in the plan, resulting in smoother, lower-energy trajectories. Although this leads to higher RMSE, it also produces higher utility due to reduced energy consumption.

This discrepancy highlights a known limitation of rolling-horizon planners: plans are generated assuming perfect tracking of previous plans, which rarely occurs in practice. Consequently, the controller continually reacts to accumulated deviation, inadvertently producing smoother behavior.

Focusing on Scenarios 2–4, which increase the red signal delay by 2, 4, and 6 seconds, we observe the following trends: For \textbf{Method A (Optimization-based)}, $G$ rises from 1.0560 in Scenario 2 to 1.0699 in Scenario 3, and remains nearly constant at 1.0689 in Scenario 4. The flat $G$ from Scenario 3 to 4 suggests that the controller can absorb additional disturbance without proportional degradation—a hallmark of resilient performance. In contrast, \textbf{Method B (Analytical)} begins at a lower $G = 1.0320$ in Scenario 2 and stays nearly constant at 1.0319 in Scenario 3, before rising to 1.0689 in Scenario 4. This sharp increase from Scenario 3 to 4 indicates sensitivity to the most severe disturbances, reflecting a lack of dynamic re-planning in the analytical controller.

These findings reveal a clear difference in resilience profiles: the optimization-based controller exhibits bounded, gradual degradation, whereas the analytical controller maintains performance under moderate stress but degrades more sharply under heavy disturbance.

Table~\ref{tab:robust} evaluates the same two methods under \textbf{internal disturbances}—execution variability, delays, and estimation noise—while external disturbances are held constant. The robustness indicator $R$ parallels $G$ but isolates internal system imperfections. For Method A, $R$ values across Tests 1 to 3 are 1.0354, 1.0182, and 1.0082, decreasing steadily. RMSE also declines from 1.2509,m to 0.6882,m. This suggests a consistent execution pattern with limited drift and a high degree of internal robustness. Method B yields $R$ values of 1.0194, 1.0072, and 1.0156—comparable to Method A in magnitude and variation. However, RMSE fluctuates more dramatically: from 0.7839,m in Test 1 to 2.4526,m in Test 2, then back to 0.8783,m in Test 3. This inconsistency reflects the controller’s sensitivity to internal errors, likely stemming from its open-loop design and lack of feedback correction.

In summary, although both methods achieve similar average robustness scores, the optimization-based controller offers more consistent tracking and bounded degradation. These observations reinforce the value of feedback-based planning and execution in maintaining stable eco-driving performance under both external and internal uncertainty.

\section{Conclusion}\label{sec:conclusion}
This paper presents a structured evaluation framework for eco-driving strategies at signalized intersections, explicitly accounting for both internal execution uncertainty and external environmental disturbance. By introducing formal definitions of robustness and resilience, we move beyond traditional efficiency-based metrics to assess how well controllers retain performance in real-world, uncertain environments.

Through field experiments involving two embedded eco-driving controllers—one optimization-based and one analytical—we demonstrated that both methods experience performance degradation under variability, but with distinct patterns. The optimization-based controller exhibited more consistent and bounded behavior in response to both internal fluctuations and external disturbances, as indicated by smoother trends in RMSE, and smaller variations in the robustness $R$ and resilience $G$ indicators. In contrast, the analytical controller maintained reasonable performance under nominal or moderately disturbed conditions but showed greater sensitivity to severe delays and execution deviations.

These results highlight the importance of incorporating robustness and resilience metrics into the evaluation and design of eco-driving strategies. The proposed framework enables more informed comparisons and supports the development of controllers that are not only energy-efficient in theory, but also reliable and adaptive in practice. Future work may extend this approach to multi-vehicle settings and explore predictive modeling of disturbance distributions for adaptive controller tuning.

\bibliographystyle{unsrt}  

\bibliography{references}
\end{document}